\documentclass{article}
\usepackage{spconf,amsmath,graphicx}
\usepackage[utf8]{inputenc}
\usepackage[spanish]{babel}

\usepackage[noend]{algpseudocode}
\usepackage{algorithmicx,algorithm}

\title{PREDICTION STABILITY AS A CRITERION IN ACTIVE LEARNING}
\name{Junyu Liu$^{1}\sthanks{Junyu Liu performed this work while at Hikvision Research Institude.}$ \qquad Xiang Li$^{2}$ \qquad Jin Wang$^{2}$ 
\qquad Jiqiang Zhou$^{2}$ \qquad Jianxiong Shen$^{3}$}
\address{$^{1}$ Graduate School of Informatics, Kyoto University, Japan \\
$^{2}$ Hikvision Research Institute, Hangzhou, China \\
$^{3}$ Institut de Robòtica i Informàtica Industrial, CSIC-UPC, Spain}

\begin{document}
%
\maketitle
\begin{abstract}
Recent breakthroughs made by deep learning rely heavily on large number of annotated samples. To overcome this shortcoming, active learning is a possible solution. Beside the previous active learning algorithms that only adopted information after training, we propose a new class of method based on the information during training, named sequential-based method. An specific criterion of active learning called prediction stability is proposed to prove the feasibility of sequential-based methods. Experiments are made on CIFAR-10 and CIFAR-100, and the results indicates that prediction stability is effective and works well on fewer-labeled datasets. Prediction stability reaches the accuracy of traditional acquisition functions like entropy on CIFAR-10, and notably outperforms them on CIFAR-100.
\end{abstract}
\section{Introduction}
\label{sec:intro}

Recent breakthroughs made by deep learning heavily rely on Supervised Learning (SL) with large amount of annotated datasets \cite{law2018cornernet,liu2018original}. But in the practical applications, large amount of labels are expensive and time-consuming \cite{qu2018orienet}. Lack of labels is an important obstacle to adopt SL methods. To achieve similar accuracy to SL with less labels, (pool-based) active learning (AL) \cite{lewis1994sequential} has become a possible solution. These strategies have succeeded in many realms such as image processing \cite{zhou2017fine} and natural language processing(NLP) \cite{tanguy2016natural}.

The goal for active learning is to select the least number of typical samples, and train the model to reach the same accuracy as one trained on all the samples. Most of the previous works select samples once after a whole training process on the existing labeled dataset. It's not difficult to find out that the core of active learning methods is the strategy of sample selection, called acquisition function. Basing on the learning process of pool-based active learning, the samples selected are expected to be the ones with most information. In many works, the selected samples are the most uncertain ones. The basic ideas include using confidence, max-entropy \cite{shannon1948mathematical}, mutual information  \cite{houlsby2011bayesian}, mean standard deviation \cite{kampffmeyer2016semantic} or variation-ratio \cite{freeman1965elementary} of samples as a measurement. Recent works of AL adopted strategies based on Bayesian Convolutional Neural Networks \cite{gal2017deep} and Generative Adversarial Nets (GAN) \cite{zhu2017generative}. Although the principle of networks is different from typical classification convolutional neural networks (CNN), the methods still generate or choose samples with highest uncertainty. There are another family of methods focuses on multi-outputs to stabilize the inputs of acquisition functions, but the thought behind the methods are still uncertainty-based. Another class of work select samples by the expectation of model change. For instance, expected gradient length \cite{settles2008analysis} choose samples expected to cause largest gradients to current model. After approximation of the algorithm, the selected samples are similar to adversarial examples \cite{Goodfellow2014Explaining}. There are also works concentrates on exploring the typical samples of the whole dataset. For example, core-set \cite{sener2017active} choose samples that are at the center of a neighbor area, and expect all the selected samples to cover the whole feature space.

Present active learning methods are different in strategy and implementation, but we can classify all the methods mentioned above as \textit{spatial-based} ones. That is, although different methods concentrate on different parts of the AL process (prediction, model updating, etc.), the information took in to account all comes from the prediction of the well-trained models before selection. The whole process is a flat one without information from the time course. Here we propose sequential-based methods, and as a verification of it, we propose a new criterion of sample selection in AL called the \textit{prediction stability}, which describes the oscillation of predictions across the epochs. Instead of starting from a well-trained model, we begin the selection process during training the model. We assume that the violent fluctuation of prediction on a sample during training means the fitting ability of model is weak in the feature area of this sample. The results of our experiments agree with our assumption and proves the proposed method as an effective one.

The following parts of this paper is divided into 4 sections. The second and third sections introduce the relation to prior work, and our methodology. The forth section provides the experimental results. And the final part is the conclusion.

\section{RELATION TO PRIOR WORK}
\label{sec:prior}

When comparing our proposed method with present AL algorithms, there are two major differences. First, our sequential-based method not only extracts features after training, but during the training process. Second, the previously proposed measures of amount of information are based on more apparent criteria including uncertainty, the influence on model and looking for typical samples. They care more about the scale of features, but prediction stability is a new criterion to catch the indirect information of relative prediction changes.

\section{METHODOLOGY}
\label{sec:method}

We can define the dataset of all samples as $X=\{x_i|i=1...n\}$, with $X^L\subseteq X$ representing the labeled set containing $n_l$ labels, and $X^U=C_{X}X^L$ is the set of unlabeled $n_u$ samples. The budget of AL is defined as $B$. For pool-based active learning, after initialization, in each round of AL, the model will select $b$ samples from $X_U$ for annotation and put the set of them $S\subseteq X^U$ into $X^L$, then the model is retrained on the new $X^L$ set. In previous works, the acquisition functions can be concluded as (1). In this equation, $f(\cdot)$ is the feature extracting function, and $g(\cdot)$ outputs the scores of samples.
$$S=\mathop{argmax}_{S}[\sum_{i=1}^{b}g(f(s_i))], s_i\in S \eqno{(1)}$$

The past spatial-based methods concentrate on the quality of final predictions. All the innovations focus on the measurements of the final prediction. Different from this kind of methods, we propose sequential-based methods that make use of the information during training. Defining number of epochs in training as $N_e$, and $f_{n}(\cdot)$ as the $f(\cdot)$ function in n-th epoch, the acquisition function can be rewritten as eq.2.
$$S=\mathop{argmax}_{S}[\sum_{i=1}^{b}g(f_{1}(s_i),..., f_{N_e}(s_i))], s_i\in S \eqno{(2)}$$

\begin{figure}[htb]
\begin{minipage}[b]{1.0\linewidth}
  \centering
  \centerline{\includegraphics[width=7cm]{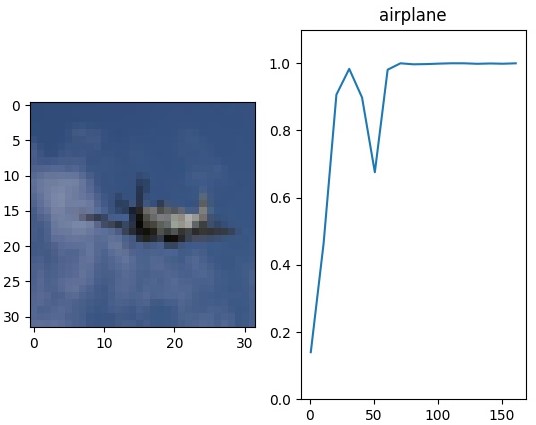}}
  \centerline{(a) Sample with high prediction stability}\medskip
\end{minipage}
\begin{minipage}[b]{1.0\linewidth}
  \centering
  \centerline{\includegraphics[width=7cm]{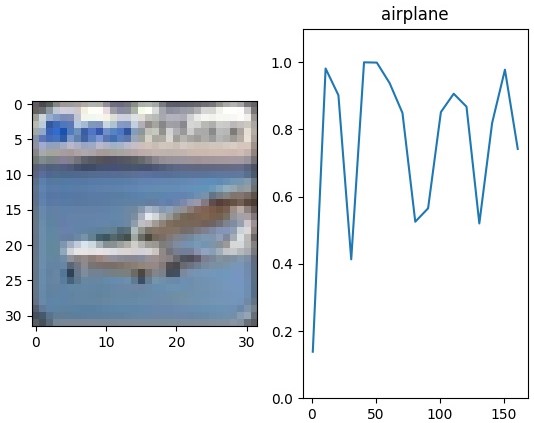}}
  \centerline{(b) Sample with low prediction stability}\medskip
\end{minipage}

%
\caption{Example of samples with different prediction stability during training. The horizontal axes in the right diagrams are index of epochs, and the vertical axes shows the scale of an element of the output vector.}
\label{fig:meth}
\end{figure}

As an application of sequential-based methods, we propose \textit{prediction stability}, a new criterion of selecting the subset $S$ in active learning. For implementation, we also adopt the common CNN model as the feature extractor and classifier.  An important distinction with former spatial-based methods is that, this criterion focus not on the real scales of feature vectors, but the fluctuation of scales during training. As Fig.\ref{fig:meth} shows, looking through the whole training process, features of samples like (a) tend to be relatively stable, but other samples like (b) oscillates from the beginning to the end. An instinct speculation is that samples like Fig.\ref{fig:meth} (b) should be selected for labeling. In order to do quantitative analysis, we test some common-used measures of fluctuation of data, and choose variance of feature vectors of different epochs as the measure of prediction stability. The diagrams in Fig.\ref{fig:meth} also shows that, due to under-fitting, the former epochs of training are definitely to violate severely. Therefore only epochs in the later training process should be included in the calculation. After experiment, we find that the selected epochs are actually at the over-fitting area, which is relatively stable. Also, considering the time complexity, only several epochs are chosen in the end. 

The definition of prediction stability can be written as eq.3:
$$g(x)=\sum_{c=1}^{C}var(f_{e_1}(x)_{c},...,f_{e_n}(x)_{c}) \eqno{(3)}$$
Where $C$ is the length of predicted feature vector by $f(\cdot)$, $f(x)_{c}$ is the c-th element of feature vector $f(x)$, and $E=\{e_1,e_2, ...,e_n\}$ is the set of index of selected epochs. The choice of $E$ is discussed in the experiment part. The whole framework is displayed in Algorithm\ref{alg:Framwork}.

\begin{algorithm}[htb] 
\caption{ Prediction Stability} 
\label{alg:Framwork} 
\begin{algorithmic}[1]
\floatname{algorithm}{Procedure}
\renewcommand{\algorithmicrequire}{\textbf{Input:}}
\renewcommand{\algorithmicensure}{\textbf{Output:}}
\Require
CNN model $M$, dataset $X=\{x_i|i=1...n\}$, initial sampling number $k$, number of epochs per training process $N_e$, set of index of selected epochs $E$, budget $B$, subset of samples selected each round $S$.

\State Generate first $k$ samples randomly, and produce labels for them;
\Repeat
    \For {$i=1\to N_e$}
        \State Train the model $M$ on labeled samples; 
        \If {$i\in E$}
            \State Predict outputs $P_i$ of $M$ on unlabeled set.
        \EndIf
    \EndFor
    \State Get prediction stability of each image along selected epochs;
    \State Select top $|S|$ samples with lowest prediction stability, generate labels and put them into labeled sample pool;
\Until Reach the budget $B$
 
\end{algorithmic} 
\end{algorithm}

\section{EXPERIMENTAL RESULTS}
\label{sec:result}

\subsection{Implementation Details}
\label{ssec:details} 

\subsubsection{Datasets}
\label{sssec:dataset}
CIFAR-10 and CIFAR-100 \cite{krizhevsky2009learning} are used for the evaluation of our proposed method. The samples of the two datasets are all $32\times32$ small image patches. Each dataset contains 50000 training samples and 10000 testing samples respectively. The training and testing samples are equally distributed into all categories. But the difference is that CIFAR-10 only has 10 classes, and CIFAR-100 contains 100 classes. Therefore, sample size in each class of CIFAR-10 is 10 times of that of CIFAR-100. 

\subsubsection{Architecture details}
\label{sssec:hyper}

As for the model $M$ for feature extraction, we employ ResNet-18 \cite{He_2016_CVPR}, which is a relatively deep architecture, and a popular choice among recent works on AL. Basically, this network mainly consists of the first convolution layer and the following 4 residual blocks. The implementation is based on an open source  framework\footnote{https://github.com/bearpaw/pytorch-classification.git}. The softmax output of network, which is the final score vector of categories, is chosen as the feature vector in this work.

All the models in this work are implemented on a NVIDIA TITAN Xp GPU.  During training, the batch size is 128, and 164 epochs are 
utilized in each training process. 
In our experiments, for each dataset, a subset containing 1000 samples is selected for the first training process. Since biases of number among different classes in the initial labeled dataset may heavily influence the selection after the first training process, equal number of samples are randomly selected from each class of the dataset in the beginning. 1000 samples are selected and labeled after each training process, and the final size of labeled dataset is 10000. To overcome the influence of random factors and get objective results, we generate 10 sets of labeled samples at first, and do the first training processes of all the methods on the same 10 datasets. Final results are the average of the ten.

\subsection{CIFAR-10}
\label{ssec:cifar10} 

\begin{figure}[htb]
\begin{minipage}[b]{1.0\linewidth}
  \centering
  \centerline{\includegraphics[width=7.5cm]{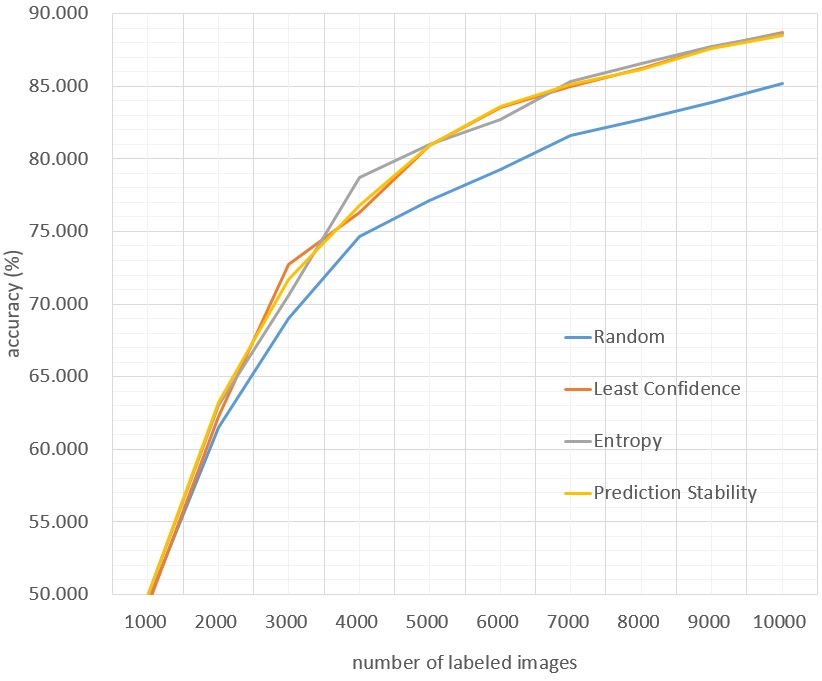}}
\end{minipage}
\caption{Results on CIFAR-10.}
\label{fig:cifar10}
\end{figure}

The results on CIFAR-10 is displayed by Fig.\ref{fig:cifar10}. Because the output features are probability of all classes, entropy and least confidence measure can be calculated on the outputs directly. For the calculation of prediction stability, we finally select 5 epochs with an interval of 5.
$$e_i=N_e-(i-1)\times interval, i=1,2,3,4,5 \eqno{(4)}$$
The results show that although information about the value of outputs are not included directly, the proposed prediction stability method still overwhelms random selection, and has similar performance with acquisition functions like entropy and least confidence on CIFAR-10. 

\subsection{CIFAR-100}
\label{ssec:cifar100}

\begin{figure}[htb]
\begin{minipage}[b]{1.0\linewidth}
  \centering
  \centerline{\includegraphics[width=7.5cm]{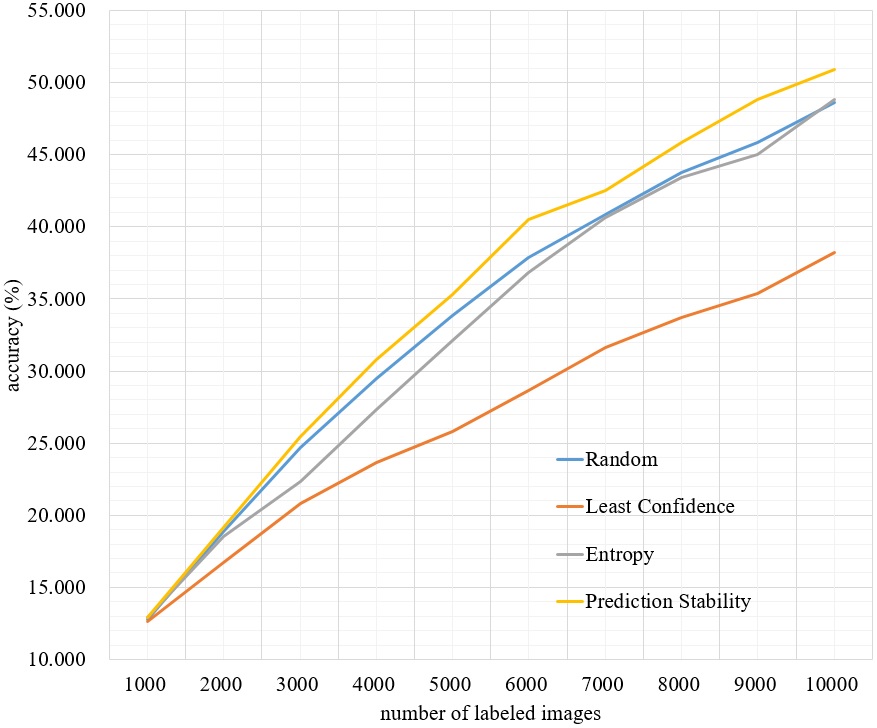}}
\end{minipage}
\caption{Results on CIFAR-100.}
\label{fig:cifar100}
\end{figure}

The performance of each method on CIFAR-100 is exhibited in Fig.\ref{fig:cifar100}. To perform prediction stability on CIFAR-100, the interval of epoch selection is set to 1. Previous works hardly report on this dataset, but our results on CIFAR-100 show totally different tendency with CIFAR-10. Entropy and least confidence, especially least confidence, suffer from deterioration of performance. Accuracy of both acquisition functions are lower than random selection. But our proposed method proves better performance and clearly outperforms random selection. 

We believe the better performance on CIFAR-100 than CIFAR-10 is caused by sample size in the feature space. The major difference between the two datasets is CIFAR-100 has less samples in each class, which means the feature space of each class is more sparse and has less labels to distinguish the boarder. The result on the two datasets proves that prediction stability has better capacity on fewer-labeled dataset.

\subsection{Ablation Study}
\label{sec:Ablation}

\subsubsection{measure of prediction stability}
\label{ssec:measure}

\begin{figure}[htb]
\begin{minipage}[b]{1.0\linewidth}
  \centering
  \centerline{\includegraphics[width=7.5cm]{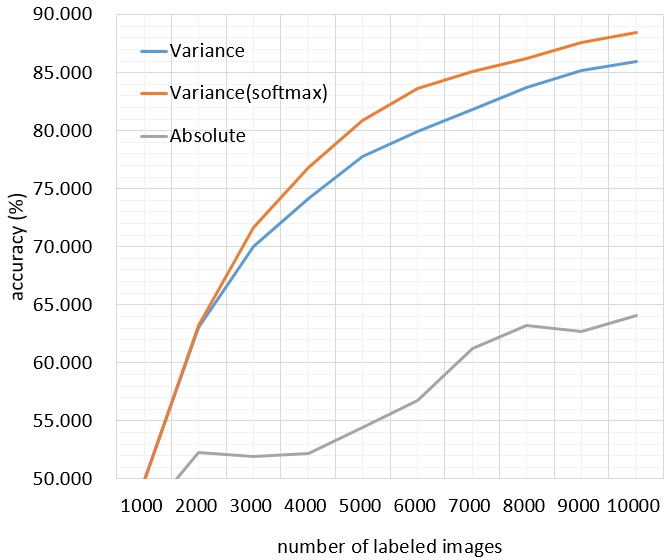}}
\end{minipage}
\caption{Results on CIFAR-10 with different measure of prediction stability.}
\label{fig:abl_measure}
\end{figure}

Experiments are made to test performance of different measure of prediction stability, as displayed in Fig.\ref{fig:abl_measure}. We test absolute increase among features of different epochs. This measure is represented by eq.5.
$$F(x)=\sum_{i=2}^{|E|}|f_{e_i}(x)-f_{e_{i-1}}(x)| \eqno{(5)}$$
The result shows that absolute increase lead to nearly 40\% drop of performance. We assume that it means it's not the tendency, but the distribution of output, that determines the performance of prediction stability. 
Also, we test the result of taking variance as the acquisition function, but leaving the output features not transformed by a softmax layer. An deterioration of result can also be observed clearly. We believe this is caused by softmax layer's function of normalization. The output features of different samples are transferred into comparable probabilities, and therefore the differences on absolute scales of output features don't influence the variances.

\subsubsection{interval of epoch selection}
\label{ssec:interval}

\begin{figure}[htb]
\begin{minipage}[b]{0.48\linewidth}
  \centering
  \centerline{\includegraphics[width=4cm]{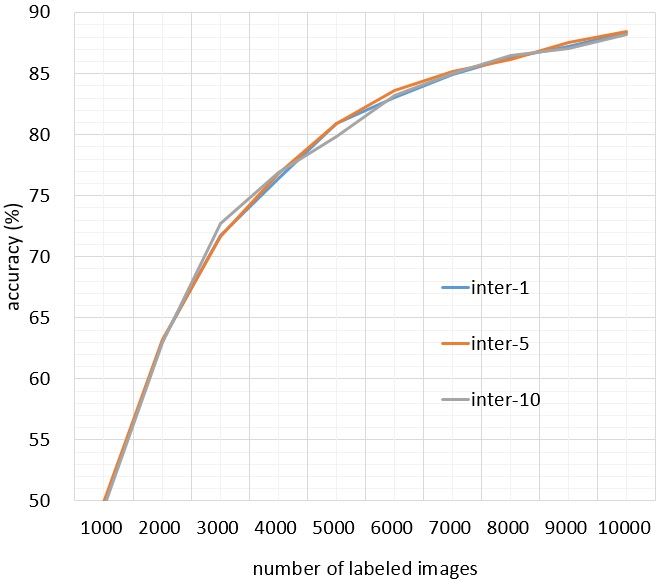}}
  \centerline{(a) Result on CIFAR-10}\medskip
\end{minipage}
\begin{minipage}[b]{0.48\linewidth}
  \centering
  \centerline{\includegraphics[width=4cm]{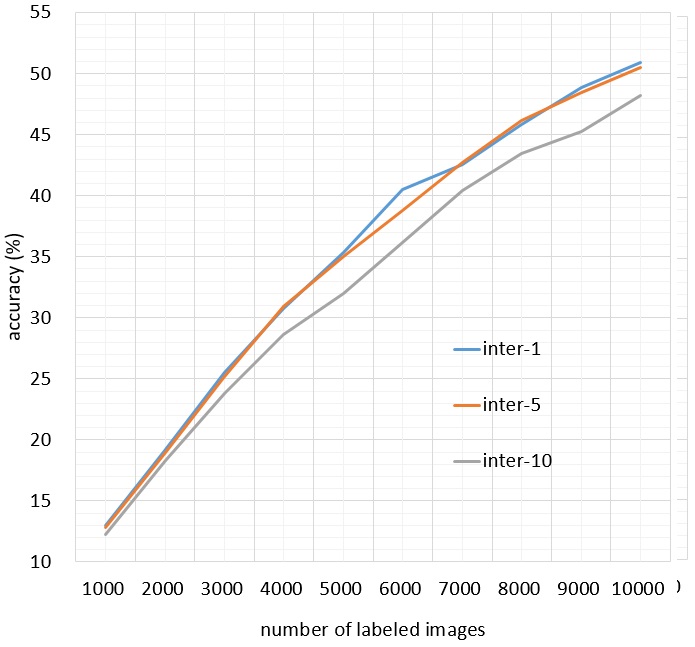}}
  \centerline{(b) Result on CIFAR-100}\medskip
\end{minipage}
\caption{Results on different intervals of prediction stability.}
\label{fig:abl_interval}
\end{figure}

Experiments are made to test influence of epoch selection on the results of prediction stability. The epoch selection process is based on eq.4. Results on the two datasets are different, as exhibited in Fig.\ref{fig:abl_interval}. Although accuracy tend to be the best when interval is 5, CIFAR-10 is not sensitive to interval change. But in CIFAR-100, the accuracy declines as interval of epoch increase. This happens may because models over-fit on CIFAR-100 later than CIFAR-10. When the interval is 10, result of some epochs of CIFAR-100 is still not stable enough and caused the decrease of accuracy.

\section{CONCLUSION}
\label{sec:conclusion}

In this paper, we propose a new class of AL method named sequential-based AL method. A new criterion, prediction stability is proposed as an application of sequential-based method. Testing results of prediction stability on CIFAR-10 and CIFAR-100 prove the feasibility of the sequential-based method class.
As for the future work, we will focus on fusing our proposed method with uncertainty-based AL methods, because the information extracted by two kinds of methods are complementary.

\vfill\pagebreak

\bibliographystyle{IEEEbib}
\bibliography{icassp}

\end{document}